\documentclass[12pt]{amsart}

\usepackage{amssymb,amscd,amsthm,extarrows,verbatim,amsmath,color,fancyhdr,mathrsfs}
\usepackage{multirow}
\usepackage{amsaddr}
\usepackage{graphicx}
\usepackage{turnstile}
\usepackage{arydshln}
\usepackage{enumitem}
\usepackage{tikz-cd}
\usetikzlibrary{arrows}

\usepackage[letterpaper, left=2.5cm, right=2.5cm, top=2.5cm,
bottom=2.5cm,dvips]{geometry}
\usepackage{xcolor}
\usepackage{pagecolor}

\newcommand{\R}{\mathbf{R}}
\def\arrvline{\hfil\kern\arraycolsep\vline\kern-\arraycolsep\hfilneg}

\title[comparator-based NN descent]
{Empirical complexity of \\comparator-based nearest neighbor descent}

\author{Jacob D.\ Baron \hspace{.8cm} R.W.R.\ Darling}
\address{National Security Agency, Fort George G.\ Meade, MD 20755-6844, USA}
\date{\today}

\begin{document}

\maketitle

\begin{abstract}
 A Java parallel streams implementation of the $K$-nearest neighbor descent
 algorithm is presented using a natural statistical termination criterion.
 Input data consist of a set $S$ of $n$ objects of type \texttt{V}, 
 and a \texttt{Function<V, Comparator<V>>}, which enables any $x \in S$
 to decide which of $y, z \in S\setminus\{x\}$ is more similar to $x$.
 Experiments with the Kullback-Leibler divergence \texttt{Comparator}
 support the prediction that the number of rounds of 
 $K$-nearest neighbor updates need not exceed twice the diameter of the
 undirected version of a random regular out-degree $K$ digraph on $n$ vertices.
 Overall complexity was $O(n K^2 \log_K(n))$
 in the class of examples studied. When objects are sampled uniformly
 from a $d$-dimensional simplex, accuracy of the $K$-nearest neighbor 
 approximation is high up to $d = 20$, but declines in higher dimensions,
 as theory would predict.
\end{abstract}

{\small
\textbf{MSC class: } Primary: 90C35; Secondary: 06A07
}

\setcounter{tocdepth}{1}
\tableofcontents

\section{Introduction}
\subsection{Context}
Baron and Darling \cite{bar} provided a theoretical analysis of the 
\textbf{$K$-nearest neighbor descent} ($K$-NN Descent) algorithm for
$K$-nearest neighbor approximation proposed and implemented by
Dong, Charikar, and Li \cite{don}.

This sequel reports on a generic Java parallel streams implementation of
$K$-NN Descent, which was written to support a forthcoming implementation 
of the partitioned nearest neighbors local depth
algorithm \cite{pannld}. While testing this implementation, we acquired
statistical data which shed light on the performance of $K$-NN Descent,
under a new termination criterion. This brief report does not attempt comparison
of $K$-NN Descent with other algorithms, as is reported in \cite{don}.
Nor shall we outline the different approaches to $K$-nearest neighbor approximation,
briefly surveyed by Aum{\"u}ller, Bernhardsson and, Faithfull \cite{aum}.
We mention the recent competition \cite{neu}
to surpass the industry leader FAISS \cite{faiss}, 
in the case of a billion dense vectors in dimension 96 to 256, under the $\ell_2$ norm.

\subsection{Previous implementations}

A high level summary of previous $K$-NN Descent implementations is shown in Table \ref{t:knndimpl}, along with
our own in the final row.

\subsubsection{Original implementation for metrics} \label{s:batchwise}
A sophisticated OpenMP and map-reduce implementation of $K$-NN Descent is described by Dong et al \cite{don}. 
These authors employ optimizations applicable to symmetric similarity functions, and employ two
stopping criteria, both of which are different to ours. The authors
describe in detail the application to five well-studied data sets 
of sizes between 28,755 and 857,820 using
several symmetric similarity measures, and compare $K$-NN Descent with Recursive Lanczos Bisection
and Locality Sensitive Hashing.

\subsubsection{Python implementation}
McInnes created the widely-used \texttt{pynndescent} Python version \cite{pynn}, which is used in UMAP \cite{mci}.
It assumes that similarity is obtained from a metric, of which 22
examples are available in the code. This is one of 19 single-threaded
approximate $K$-NN Python algorithms among benchmarks at \cite{ber}.

\subsubsection{Single-threaded C Implementation}
Kluser et al \cite{klu} describe a runtime-optimized C implementation for the $\ell_2$-distance
metric, and report performance improvements over the two versions above. Their approach increases locality
by improving the otherwise irregular memory access pattern.

\begin{table}
\caption{
Comparison among four $K$-NN Descent implementations.
}\label{t:knndimpl}
\centering
\begin{tabular}{| l | c | c | c| c |} 
\hline 
Authors & Language & Asymmetric? & Parallelization? & Termination \\ \hline
Dong et al \cite{don} & C++ & no & OpenMP \& map-reduce & $\delta$-proportion update \\ \hline
McInnes \cite{pynn} & Python & no & none & no possible update \\ \hline
Kluser et al \cite{klu} & C & no & none & no possible update \\ \hline
This paper & Java & yes & fork-join pool & statistical criterion\\
 \hline
 \end{tabular}
\end{table}

\subsection{Novelty of our implementation}
Here are some more details, beyond Table \ref{t:knndimpl}, of what distinguishes our implementation from the others.

\begin{enumerate}
 \item[(a)] \textbf{Non-metric: }Input data consist of a set $S$ of $n$ objects of type \texttt{V}, 
 and a \texttt{Function<V, Comparator<V>>}, which enables any $x \in S$
 to decide which of $y, z \in S\setminus\{x\}$ is more similar to $x$. This ``triplet comparison'' has
 more general application than similarity induced by a symmetric distance function, as we discuss in \cite{pannld}, but
 disallows optimizations based on symmetric numerical similarity functions, used in the works cited above.
 
 \item[(b)] \textbf{Stopping criterion: }Termination depends on a statistical criterion, presented in Section \ref{s:termination},
 applied to a sampled quantity called the friend clustering rate.
 By contrast, other implementations continue until no further updates are possible, except for a variant by \cite{don}
 which stops when no more than a proportion $\delta$ of points allow updates.
 
 \item[(c)] \textbf{Parallelism: }By casting the algorithm into a functional
 programming framework, we enable the Java Virtual Machine to distribute
 tasks among threads via a fork join pool, invisible to the programmer. Speedup due to parallelism is reported below.
\end{enumerate}

\section{$K$-nearest neighbors based on \texttt{Comparator}s} \label{s:javannd}
\subsection{Setting}
Input data consist of a set $S$ of $n$ objects of type \texttt{V}, 
and what we call a \textbf{ranking system} \cite{bar}, which
attaches to each $x \in S$ a total order\footnote{By definition, 
this relation is anti-symmetric and transitive for all $x$.} $\prec_x$ on $S \setminus \{x\}$; here
$y \prec_x z$ is interpreted to mean that $y$ is more similar to $x$ than $z$ is.
In data science, this is called \textit{triplet comparison}.
The computer science equivalent is a \texttt{Function<V, Comparator<V>>}. Here we map each $x \in S$ to a
specific \texttt{Comparator} \cite{java}, whose \texttt{compare} method depends on $x$. Formally
\texttt{compare}$(x; y, z) < 0$ means $y \prec_x z$, and \texttt{compare}$(x; y, z) > 0$ means
$z \prec_x y$, for distinct $x, y, z \in S$.

Such a family of \texttt{Comparator}s gives an orientation\footnote{
An orientation of a graph $G$ is a digraph
obtained by replacing each edge $\{a, b\}$ by one
of the arcs $a \rightarrow b$ or $b \rightarrow a$. The textbook \cite{ban} explains graph-theoretic terms.
} of the line graph\footnote{
Undirected edges of the complete graph $K_n$ on $S$ form the points of
the line graph $\mathcal{L}(K_n)$, in which $xy \sim xz$ for distinct $x, y, z$.
Here $xy$ is an abbreviation for the edge $\{x,y\}$ of $K_n$.
} $\mathcal{L}(K_n)$,
called the \textit{ranking digraph}. The relation $y \prec_x z$ is interpreted as
an arc $xy \rightarrow xz$. 
See Figure \ref{f:dag15} for an example.
No comparison between $xy$ and $zw$ is provided if $\{x,y,z,w\}$
is a set of size four.
 
 Most authors study the special case of a metric $\rho$ on $S$, where
$y \prec_x z$ means $\rho(x, y) < \rho(x, z)$.
Thus a metric (without ties) gives a total order (by distance)
on points of the line graph $\mathcal{L}(K_n)$.

\textbf{Theorem:} (Baron and Darling \cite[Lemma 5.5]{bar}) \textit{The ranking digraph is acyclic
if and only if the \texttt{Comparator}s arise from some metric.}

\subsection{Example of non-metrizable \texttt{Comparator}}
\begin{figure}
\caption{\textit{Ranking digraph on $\binom{6}{2}$ vertices,
forming the unordered pairs from a set $\{x_i\}_{1 \leq i \leq 6}$, with each $x_i$
drawn uniformly at random from a
14-simplex. Vertex label $\{1, 6\}$ (bottom right) refers to the unordered pair $\{x_1, x_6\}$.
Its neighbors are $\{1, 2\}, \{1, 3\}, \{1, 4\}, \{1, 5\}$ together with 
$\{2, 6\}, \{3, 6\}, \{4, 6\}, \{5, 6\}$. Orientation of an arc such as 
$\{1,2\}\rightarrow \{1,4\}$ means that Kullback-Leibler divergences satisfy
$D(x_1 \| x_2) < D(x_1 \| x_4)$. Proof by counterexample that this 
ranking system is not metrizable: vertices marked in black form a cycle 
\[
(\{1,2\}\rightarrow \{1,4\},\{1,4\}\rightarrow\{2,4\},	\{2,4\}\rightarrow\{1,2\}).
\]
}}
\label{f:dag15}
\begin{center}
\scalebox{0.33}{\includegraphics{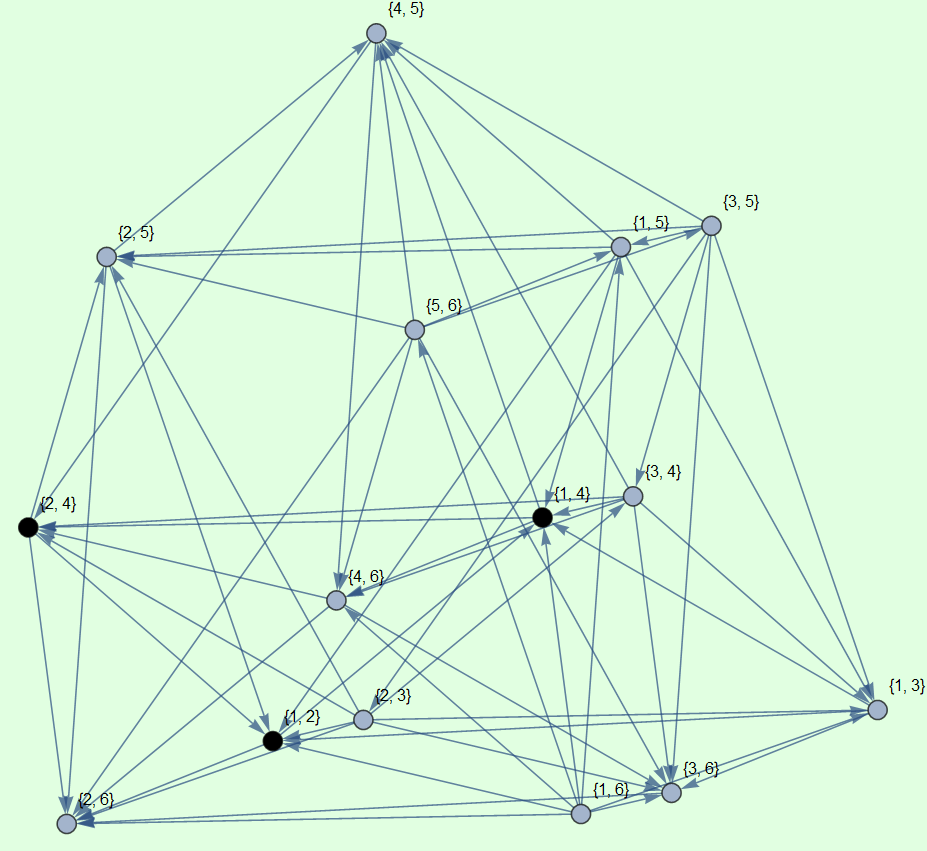} }
\end{center}
\end{figure}

For points $x, y, z$ in the interior of the $d$-dimensional simplex (here $d \geq 3$),
choose this \texttt{Comparator}: $y$ is closer to $x$ than $z$ is when
\[
D(x \| y) < D(x \| z), 
\]
where $D(x \| y):=\sum_1^d x_i \log{(x_i/y_i)}$ is Kullback-Leibler (KL) divergence.
KL divergence is asymmetric in the pair $(x, y)$, and does not satisfy the
triangle inequality. This example was chosen to dispel the notion that
a metric is needed.
An example of a ranking digraph from six randomly generated points
is shown in Figure \ref{f:dag15}.
The existence of a 3-cycle shows that these \texttt{Comparator}s are not metrizable.
Further examples and properties of ranking digraphs are described in \cite{pannld}.

\subsection{$K$-NN graph}
The $K$-NN graph is the \textbf{directed graph}
with an arc from each $x \in S$ to the
$K$ elements of $S \setminus \{x\}$
most similar to $x$. Naive computation would take $n$ calls to a 
\texttt{Comparator}-based sorting operation on $n-1$ objects,
which would be $O(n^2 \log{n})$ work -- or in the special case of a metric,
$\binom{n}{2}$ distance evaluations followed by an $O(n^2 \log{n})$ sort.
An ideal outcome would be to approximate the $K$-NN graph using $O(n \log_K{n})$ 
calls to a \texttt{Comparator}-based sort of $O(K^2)$ objects.

Before we describe how $K$-NN Descent works, we mention that \cite{pannld} lists 
half a dozen data science algorithms besides
$K$-NN Descent that accept triplet comparisons as input, 
including another nearest neighbor search algorithm \cite{hag}.

\section{$K$-NN descent algorithm with a statistical stopping rule}
\subsection{Friend-of-a-Friend Principle}
Dong, Charikar, and Li \cite{don} base $K$-NN Descent on the
\textit{Friend-of-a-Friend Principle},
which states that \textit{a friend of a friend may be suitable as a friend}.
The algorithm consists of a sequence of rounds.
At each round, we have a digraph on $S$, with regular out-degree $K$. Such a graph
is called a $K$-\textbf{out} graph \cite[Ch. 16]{fri}.
If $x \rightarrow y$, call $y$ a \textbf{friend} of $x$, and $x$ a \textbf{co-friend} of $y$.
If there is a ranking system on $S$, one $K$-out graph
may be replaced by another $K$-out graph using a procedure we call
\textbf{friend set update}.

Friend set update is like a cocktail party, attended by all members of $S$.
\begin{enumerate}

\item[Meet:]
Each $x$ ``meets'' all friends of friends, and friends
of co-friends, which become new acquaintances.

\item[Update:]
The new friend set of $x$ consists of the closest $K$ out of all 
new acquaintances and former friends.
\end{enumerate}

\subsection{Initialization of $K$-NN Descent}
\begin{itemize}
\item
Each $x$ selects $K$ elements of $S \setminus \{x\}$
uniformly at random as its initial friend set. This digraph is called
\textit{random $K$-out} (Frieze \& Karonski \cite[Ch. 16]{fri}).
\item
Random $K$-out is an expander graph, whose
undirected version has diameter
$\leq \lceil \log_{K-1}{n} \rceil$ with high probability \cite[Appendix]{bar}.
\item
\textbf{Plausible heuristic: } $2 \lceil \log_{K-1}{n} \rceil$ rounds
of cocktail parties (friend updates) should suffice for
``everyone to get to know each other''. Twice the diameter allows a ``message'' to
travel from any vertex to its antipode, and back.
\end{itemize}

\subsection{Sampling Method for Termination for $K$-NN Descent} \label{s:termination}

\begin{enumerate}
\item[(i)]
Sample uniformly a point $x \in S$, and two friends $y, z \in S$ of $x$. 
\item[(ii)]
The \textit{friend clustering rate} is
 the sample relative frequency that $y$
is a friend or co-friend of $z$.
\item[(iii)]
The friend clustering rate is close to zero at the outset, and increases during the algorithm towards a plateau.
\item[(iv)]
Stop at the first round at which the friend clustering rate
does not increase, compared to the previous round.
\end{enumerate}

\section{A Java implementation of $K$-NN descent}
\subsection{Java parallel streams}

The tools of modern Java \cite{java}, including generic types, pure functions,
and parallel streams, enable a concise and performant distributed
implementation of NND. 
Four main Java types are listed in the left column of
Table \ref{t:javatypes} in monospaced font, and instances of these types 
used in NND are listed in the right column in
boldface font.

\begin{table}\caption{Ingredients of the Java functional implementation of NND.
For example, the Java object \textbf{rankingSystem} is an instance of the Java type
 \texttt{Function}$\langle \langle V$, \texttt{Comparator}$\langle V \rangle \rangle$.
}\label{t:javatypes}
\centering
 \begin{tabular}{||c c ||} 
 \hline
 Java Type & Instance Object\\ [0.5ex] 
 \hline\hline
 \texttt{List}$\langle V \rangle$ & \textbf{points} \\ 
 \texttt{Function}$\langle \langle V$, \texttt{Comparator}$\langle V \rangle \rangle$ & \textbf{rankingSystem} \\
 \texttt{Map}$\langle V$, \texttt{Set}$\langle V \rangle \rangle$ & \textbf{friends, coFriends} \\
\texttt{Function}$\langle V$, \texttt{Set}$\langle V \rangle \rangle$ & \textbf{proposeNewFriendSet}\\ [1ex] 
 \hline
 \end{tabular}
\end{table}

Elements of $S$ have a generic type \texttt{V}, supplied 
by the invoking class (e.g. strings, vectors, trajectories). 
A ranking system is a \texttt{Function}
from \texttt{V} to a \texttt{Comparator} of objects
of type \texttt{V}.
Given three elements $x, y, z$ of $S$, the assertion that
$y \prec_x z$ is equivalent in Java to:
\begin{center}
\textbf{rankingSystem}.apply$(x)$.compare$(y, z) < 0$.
\end{center}
The \texttt{Comparator} for object $x$ is visible as
 an $x$.getComparator() method of the class \texttt{V}.

\subsection{Distribution of tasks to processors}
Initialize the \textbf{friends} \texttt{Map} so that the value associated with the key $x$ is
a \texttt{Set}$\langle V \rangle$ object containing
$K$ elements of $S \setminus \{x\}$ selected uniformly at random.
The \textbf{coFriends} \texttt{Map} is derived from the \textbf{friends} \texttt{Map}.

Given $x$, the function \textbf{proposeNewFriendSet} 
(bottom of Table \ref{t:javatypes}) compares
all the cofriends, friends of friends, and friends of cofriends, with
the current friend set of $x$, and proposes the best $K$ of all these
 as a new friend set. It is crucial that \textit{the friends of
 $x$ are not updated at the time the function is called}. The \textbf{friends} and \textbf{coFriends}
 Maps remain \textit{effectively immutable} while all these proposals are constructed in
 a parallel stream. This is part of the contract of \texttt{java.util.stream}, 
 and makes it possible to execute a round of NND in a single line of code,
by invoking the \texttt{collect()} method of \texttt{Stream} \cite{java}:

\begin{center}
\begin{small}
 \textbf{points}.parallelStream().collect(Collectors.toMap($x \to x, x \to$
 proposeNewFriendSet.apply$(x)$)).
\end{small}
\end{center}
The \texttt{Map}$\langle V$, \texttt{Set}$\langle V \rangle \rangle$ produced by this command becomes the \textbf{friends} \texttt{Map} for the next round,
and \textbf{coFriends} is updated accordingly.

The value type of the \textbf{friends} \texttt{Map} is
\texttt{NavigableSet}$\langle V \rangle$, with respect to the \texttt{Comparator}.
Initial friends are ordered on insertion.
During a friend set update at $x$, each new candidate is compared to
the last member of the \texttt{NavigableSet} at $x$, and replaces it
if appropriate.

The Java Virtual Machine allocates
\textbf{proposeNewFriendSet} tasks among the processors and threads
at run time. For example, experiments on a 12-core workstation 
with JRE 11
gave an eightfold speedup\footnote{
Here $n = 2 \times 10^6$, $K=16$. The speedup was much less for smaller $n$.
}, per round, compared to the same
code where a single stream was used instead of a parallel stream.
With $n$ points and $p$ processors, the speedup should be
monotonically increasing in $\frac{n}{p}$ for fixed $K$, assuming that
the limitation is thread contention for access to the \textbf{friends} and
\textbf{coFriends} maps.

\begin{table}
\caption{
Scaling observed in parallel streaming NND
(dual Intel X5660, 12 cores total, JRE 11).
Processing times decrease in successive rounds,
because duplicate candidates are proposed by different friends.
 Time to
execute a single round of NND scaled linearly with $n$.
When $K$ was doubled, time to execute a round increased by a factor less than four.
The FCC column shows the final value of the
friend clustering coefficient, which appears to decrease with $n$.
The proportion of the true
$K$ nearest neighbors found, among a uniform sample of six points,
was typically $95\%$ or better; see Table \ref{t:dimension}. Results are consistent with the heuristic that 
$2 \lceil \log_K{n} \rceil$ rounds of NND suffice.
}\label{t:scaling}
\centering
 \begin{tabular}{| l | c | c | c | c | c | c |}
 \hline
 $n$ & $K$ & rounds & $2 \lceil \log_K{n} \rceil$ & 1st round & last round & FCC \\ [0.5ex]
 \hline
 \hline
$2 \times 10^4$ & 16 & 5 & 8 & 1.9 sec & 0.35 sec & 0.271 \\
$\,$ & 32 & 6 & 6 & 2.6 sec & 0.9 sec & 0.264 \\
 \hline
 $2 \times 10^5$ & 16 & 7 & 10 & 9.1 sec & 5.3 sec & 0.210 \\
 $\,$ & 32 & 5 & 8 & 26 sec & 13 sec & 0.231 \\
 \hline
 $2 \times 10^6$ & 16 & 8 & 12 & 88 sec & 56 sec & 0.205 \\
 $\,$ & 32 & 7 & 10 & 295 sec & 153 sec & 0.210 \\
 $\,$ & 64 & 6 & 8 & 1059 sec & 401 sec & 0.215 \\
 \hline
 \end{tabular}
\end{table}

\subsection{How many rounds of NND are needed? }
How many rounds of friend updates do we expect
before the termination criterion of Section \ref{s:termination},
is satisfied?
Baron and Darling \cite{bar}, and other citations therein, justify
$\lceil \log_K{n} \rceil$ as an estimate
for the diameter of the undirected graph on $S$ whose edges
are the pairs $\{x, y\}$, where $y$ is an initial friend of $x$.
In our experiments, the number of rounds never exceeded, but was close to,
$2 \lceil \log_K{n} \rceil$; see Table \ref{t:scaling}.

\subsection{Scaling of execution time with $n$ and $K$: }
The type \texttt{V} that we chose for our timing experiments was
a point on the interior of the
9-dimensional standard simplex in $\R^{10}$,
representing a probability measure on a set of size ten.
The ranking system was defined by taking
$y \prec_x z$ whenever
\[
D(x \| y) < D(x \| z)
\]
where $D(x \| y)$ denotes Kullback-Leibler divergence of $y$
from $x$.The points themselves were sampled
from a 10-dimensional Dirichlet distribution.
Results are shown in Table \ref{t:scaling} and discussed
in the caption. The practical implications are:
\begin{enumerate}
\item A single call to \textbf{proposeNewFriendSet} costs
 $O(K^2 \log{K})$ work on average\footnote{On average
$K + 2 K^2$ items or fewer are proposed for insertion into a sorted
set of size $K$.}. Each round needs $n$ calls to \textbf{proposeNewFriendSet}.

 \item Run time for a single round of NND scales linearly with $n$,
 for fixed $K$.
 
 \item The number of rounds of NND is not observed to exceed
 $2 \lceil \log_K{n} \rceil$, suggesting an overall
 $O((n \log{n} ) K^2)$ run time, on cancelling two $\log{K}$ factors.
 
\item For points on a 9-dimensional simplex, the accuracy (or recall)
is 95\% or better using our chosen stopping criterion, where accuracy means
proportion of the true
$K$ nearest neighbors found, among a uniform sample of six\footnote{
We did not choose a larger sample size than six, because the random initialization
already causes random variation in the outcome when NND is applied
repeatedly to the same data set.
} points.

 \item On $p$ processors, parallel streams yield a speedup 
 slightly less than $p$, presumably because of thread contention.
 We observed at best an 8 times speedup on 12 cores.
 
\end{enumerate}

\begin{table}
\caption{\textit{
Here $n = 2 \times 10^5$ points were sampled from the 
$(d-1)$-dimensional simplex in $\R^d$, for four different values of
$d$, and $K$-NN descent was performed for $K = 16$ and $K = 64$.
All runs finished within $2 \lceil \log_K{n} \rceil$ (8 or 6)
rounds.
The FCC row shows the final value of the
friend clustering coefficient.
The accuracy row shows proportion of the true
$K$ nearest neighbors found, among a sample of six points.
Note the gradual decline of accuracy with
dimension, especially when the dimension $d-1$ exceeds the
number $K$ of neighbors. }
}\label{t:dimension}
\centering
 \begin{tabular}{| l| c | c | c | c | c |} 
 \hline
$K$ & feature & $d = 10$ & $d = 20$ & $d = 40$ & $d = 60$ \\ [0.5ex]
 \hline
 \hline
64 & FCC & 0.24 & 0.13 & 0.08 & 0.06 \\ 
$\,$ & accuracy & 1.0 & 1.0 & 0.90 & 0.84 \\ \hline
 16 & FCC & 0.21 & 0.13 & 0.09 & 0.08 \\ 
$\,$ & accuracy & 0.95 & 0.52 & 0.43 & 0.36 \\ \hline
 \end{tabular}
\end{table}

\subsection{Effect of dimension}
The experiments in Table \ref{t:scaling} were performed
on points from a 9-dimensional simplex, taking $K = 16, 32, 64$.
We also performed experiments where 
the points were drawn from $(d-1)$-dimensional simplices,
for $d = 10, 20, 40, 60$,
comparing the cases $K = 16$ and $K=64$.
Table \ref{t:dimension}
shows a decline both in the friend clustering coefficient, and 
in the accuracy of the $K$-NN approximation
as dimension increases, for fixed $K$. 
Similar results are reported by Dong et al \cite[Section 4.5]{don},
who interpret them as a consequence of the fact that, when sampling many points at random
in high dimensions, the nearest neighbor and farthest neighbor of any point
are at roughly the same distance \cite{bey}.

\section{Conclusions and future work}
In the benign setting of probability measures sampled uniformly at random from
a simplex in Euclidean space, with a comparator based on Kullback-Leibler divergence,
performance of $K$-nearest neighbor descent conforms to the predictions 
based loosely on expander graphs. In particular, our statistical stopping criterion
is satisfied within $2 \lceil \log_K{n} \rceil$ rounds on a set of $n$ points,
giving a run time proportional to
\begin{equation}
 K^2 n \log{n}
\end{equation}
in contrast to the $O(n^{1.14})$ run time (for fixed $K$) reported 
by Dong et al \cite{don}. The accuracy of NND in (intrinsic) dimension up to 20
is entirely satisfactory in examples studied, and does not require that the similarity measure
be symmetric or derived from a metric.

Performance of NND can be much worse in other settings, such as a collection
of long multi-character strings under a metric based on the longest common 
substring \cite[Section 4.4]{bar}. 
More theory and new
experiments will be needed to delineate the contexts in which 
$K$-nearest neighbor descent works well.

\textbf{Acknowledgment: } The authors thank James Maissen for his comments
and suggestions on the manuscript.

\end{document}